\documentclass{article}

\usepackage[final,nonatbib]{neurips_2021}

\usepackage{bm}
\usepackage{listings}
\usepackage[dvipsnames]{xcolor}
\usepackage{cite}
\usepackage{amsmath,amssymb,amsfonts}
\usepackage{algorithmic}
\usepackage{graphicx}
\usepackage{textcomp}
\usepackage{algorithm}
\usepackage{bm}
\usepackage{amsthm}
\usepackage{amsmath}
\newtheorem{remark}{Remark}
\usepackage{subfigure}
\makeatletter
\renewcommand{\fnum@figure}{Fig. \thefigure}
\makeatother

\def\BibTeX{{\rm B\kern-.05em{\sc i\kern-.025em b}\kern-.08em
    T\kern-.1667em\lower.7ex\hbox{E}\kern-.125emX}}

\title{Safety-guaranteed trajectory planning and control based on GP estimation for unmanned surface vessels}
\usepackage[utf8]{inputenc} 
\usepackage[T1]{fontenc}    
\usepackage{hyperref}       
\usepackage{url}            
\usepackage{booktabs}       
\usepackage{amsfonts}       
\usepackage{nicefrac}       
\usepackage{microtype}      
\usepackage{xcolor}         

\author{%
    Shuhao Zhang\thanks{The main work was done while Shuhao Zhang was at the University of Melbourne. Shuhao Zhang and Yujia Yang have contributed equally to this work.}\\
    Department of Mechanical\\
    Katholieke Universiteit Leuven\\
    \texttt{shuhao.zhang@kuleuven.be}\\
   \And
    Yujia Yang$^\ast$\\
    Department of Electrical and Electronic Engineering\\
    University of Melbourne\\
    \texttt{yujyang1@student.unimelb.edu.au}\\
   \And
    Seth Siriya\\
    Department of Electrical and Electronic Engineering\\
    University of Melbourne\\
    \texttt{ssiriya@student.unimelb.edu.au}\\
   \And
    Ye Pu\\
    Department of Electrical and Electronic Engineering\\
    University of Melbourne\\
    \texttt{ye.pu@unimelb.edu.au}\\
}

\begin{document}

\maketitle

\begin{abstract}
    We propose a safety-guaranteed planning and control framework for unmanned surface vessels (USVs), using Gaussian processes (GPs) to learn uncertainties. 
    The uncertainties encountered by USVs, including external disturbances and model mismatches, are potentially state-dependent, time-varying, and hard to capture with constant models.
    GP is a powerful learning-based tool that can be integrated with a model-based planning and control framework, which employs a Hamilton-Jacobi differential game formulation. Such combination yields less conservative trajectories and safety-guaranteeing control strategies.
    We demonstrate the proposed framework in simulations and experiments on a CLEARPATH Heron USV.
\end{abstract}

\section{Introduction}
USVs have emerged as powerful platforms for missions such as environmental and climate monitoring, surface surveillance, and rescuing. 
However, uncertainties like strong waves and winds pose major challenges to USV applications.
Even in calm waters, the presence of state-related uncertainties such as skin friction and wave drift \cite{b10} lead to model mismatches in USV dynamics and make safe control difficult.
Therefore, in motion planning and control tasks, it is vital to provide safety guarantees for USVs. 
Simultaneously, it is desirable to have the ability to learn and update uncertainty models to avoid conservative performance in complex environments.

Trajectory planning and control methods for marine vehicles have previously been studied using techniques such as dynamic programming \cite{b1_seth} and model predictive control (MPC) \cite{b2_seth}, but these methods lack robustness to uncertainties. 
Thus, there is an increasing interest in safety-guaranteed methods for control; some techniques include robust MPC \cite{b3_seth}, safety funnels \cite{b4_seth} and control barrier functions \cite{b5_seth}. 
A promising body of work is methods based on Hamilton-Jacobi (HJ) reachability analysis for systems under uncertainties, including FaSTrack \cite{b11}.
FaSTrack proposes a framework for real-time safety-guaranteed planning and control under bounded disturbances via a HJ differential game. 
It has been applied to autonomous underwater vehicles with wave disturbance in \cite{b12}, where the authors assume a disturbance with a pre-specified model. 
However, there are complex uncertainties for which it is difficult or sometimes impossible to find a deterministic model in real environments.
Learning-based methods offer attractive alternatives for characterizing state-dependent uncertainties. 
In \cite{b13}, GP-based estimation of unmodeled dynamics is applied to learn a narrow uncertainty bound, which is combined with a HJ-based method to generate a less conservative control policy.

\begin{figure}[htbp]
    \centering
    \begin{minipage}[t]{0.5\textwidth}
        \centering
            \includegraphics[scale=0.22]{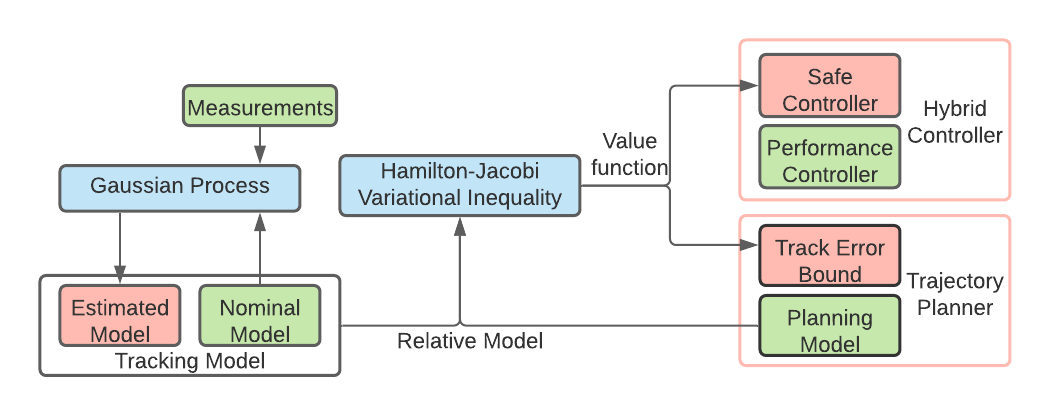}
        \caption{Offline stage of the proposed framework}
        \label{pic_1}
    \end{minipage}
    \begin{minipage}[t]{0.38\textwidth}
        \centering
            \includegraphics[scale=0.24]{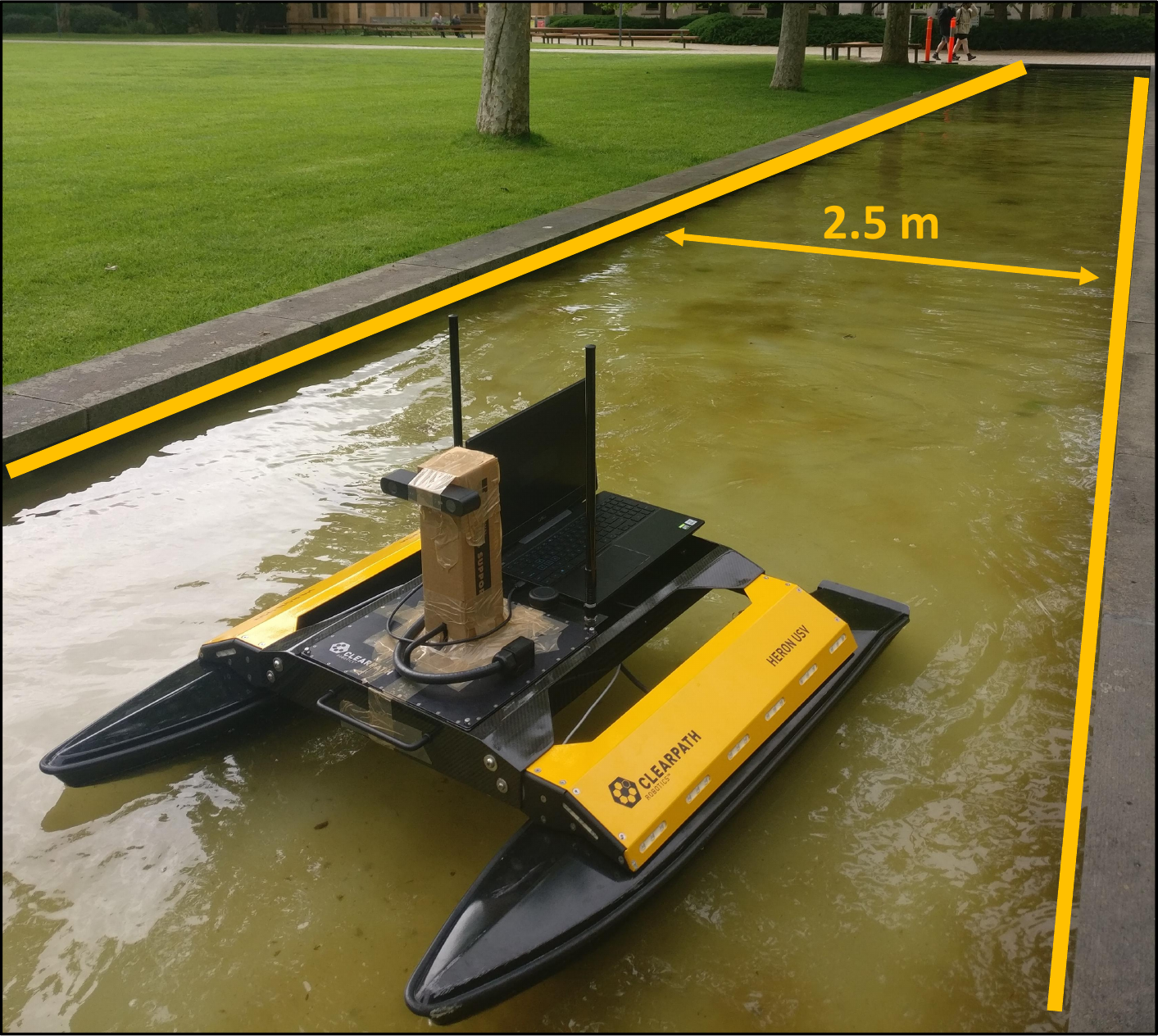}
        \caption{The Heron on a narrow tunnel}
        \label{pic_2}
    \end{minipage}
\end{figure}

In this work, we apply GPs to learn state-dependent uncertainties, allowing the model considered in the FaSTrack framework to achieve less conservative real-time planning and control, where the framework can be updated based on new measurements of the environment. 
As shown in Fig. \ref{pic_2}, we are particularly interested in driving the USV safely through a narrow tunnel, where we observe unmodeled dynamics when the boat is close to the edges. 
This phenomenon, which is uncommon to USV applications in open water, together with other model mismatches and disturbances, will be considered by our framework. 
We demonstrate the proposed framework both in simulation and real experiments using a CLEARPATH Heron USV\cite{b14}.
\section{Safety-guaranteed trajectory planning and control with GP estimation}
The FaSTrack framework \cite{b11} guarantees safety by solving a Hamilton-Jacobi differential game with the worst-case bound $\mathcal{D}$ for uncertainty descriptions.
However, for USV applications, the worst-case bound may lead to limited performance or even infeasibility in trajectory planning and control.
Therefore, we introduce the Gaussian process into the framework to learn uncertainties and update the uncertainty descriptions for reducing conservativeness while providing a chance-constrained safety guarantee. 
The proposed framework has an offline stage and an online stage.
Fig. \ref{pic_1} shows its offline stage, which has two offline computation phases; estimating uncertainties via GPs, and deriving the safe controller as well as the tracking error bound (TEB) via a HJ differential game formulation, denoted by two blue blocks, respectively. 
The pink blocks indicate the outputs of the offline computation stages, and the green blocks indicate their inputs as well as the performance controller used for real-time control.
The pink frames group the components of the hybrid controller and the trajectory planner, respectively, which can be used in the online stage.
In this section, we present the details of the offline computation phases, the corresponding model formulations, and the proposed framework with online implementation.
\subsection{The nominal model of the unmanned surface vessel}\label{sec_21}
We first present the USV's nominal model, which captures a major part of the USV dynamics without considering the uncertainties. 
Introducing a nominal model is necessary for both the estimation of underlying uncertainty models and the computation in the HJ differential game formulation. 
Based on the 5D horizontal simplified model from Fossen's equations of motion (EOMs) in \cite{b10}, the USV's \textit{nominal} model $\dot{s}=f_0(s, u_s)$ used in this work is
\begin{equation}
    \dot{s}= f_0(s, u_s)
    \triangleq 
        \begin{bmatrix}
        \Dot{x}\\
        \Dot{y}\\
        \Dot{\psi}\\
        \Dot{v}\\
        \Dot{\omega}
    \end{bmatrix} = 
    \begin{bmatrix}
        v\cos{\psi} \\
        v\sin{\psi} \\
        \omega\\
        m^{-1}((n_1+n_2)F_{max}-(X_v+X_{|v|v}|v|)v)\\
        I^{-1}((n_1-n_2) \frac{L}{2} F_{max}-(N_{\omega}+N_{|\omega|\omega}|\omega|)\omega)
    \end{bmatrix},
    \label{eq:nominal}
\end{equation}
with the left and right thruster commands as the tracking inputs $u_s=[n_1,n_2]^\top\in\mathcal{U}_s$. They are related to the forward force $F\triangleq(n_1+n_2)F_{max}$ and rotation torque $M\triangleq(n_1-n_2) \frac{L}{2} F_{max}$ in (\ref{eq:nominal}), where length is $L = 0.7366m$, and the maximal force is $F_{max} = 45N$. 
The tracking states $s\in\mathcal{S}$ of the nominal model (\ref{eq:nominal}) include the positions in global $x$ and $y$ directions, the yaw angle $\psi$, the related forward velocity $v$ between the USV and the water flow, and the yaw rate $\omega$.
Other constant parameters in (\ref{eq:nominal}) are the Heron USV's mass $m=36kg$, its inertial tensor $I=8.35kg.m^2$, quadratic damping terms $X_{|v|v}=16.9N/(m/s)^2, N_{|\omega|\omega}=13.0N.m/(rad/s)^2$ with zero linear damping terms $X_v, N_\omega$ in forward and angular velocities, respectively, which are identified by \cite{b22}.

\begin{remark} 
The inputs of the nominal model (\ref{eq:nominal}) are named by the \textit{tracking} states $s$ and the \textit{tracking} inputs $u_s$ respectively, to be consistent with \cite{b11}, because the USV's \textit{tracking} model, which is expected to capture the whole USV dynamics under the real environment, will be defined with the same states $s$ and control inputs $u_s$.
Therefore, the model mismatch between the \textit{nominal} model (\ref{eq:nominal}) and the proposed \textit{tracking} model in the following section is one kind of (state-dependent) uncertainty that we want to eliminate with learning-based methods such as the Gaussian process.
\end{remark}
\begin{remark} 
In addition to Remark 1, we point out other reasons for the necessity to introduce learning-based methods to learning uncertainties in terms of the proposed model:
1) We use the parameters identified by \cite{b22}, however, parameters may deviate in different environments, and vary depending on the payload. 
The model mismatch caused by the wrong parameters should also be dealt with.
2) State-related external disturbances like water flow are the main part of the uncertainties affecting the USV. 
Representing these uncertainties with predefined state-dependent models is impractical, which motivates the use of learning-based approaches, such as the Gaussian process introduced in the next section, to model them and subsequently account for them.
\end{remark}
\subsection{Estimating uncertainties via Gaussian processes}\label{sec_22}
We use Gaussian processes to estimate the uncertainties that cannot be captured by the nominal model (\ref{eq:nominal}).
A Gaussian process is specified by a mean function and a covariance kernel function, which inherits the advantage of the Guassian distribution,
and allows for Bayesian inference based on finite observations (e.g. measured uncertainties in this context).
We refer to \cite{b23} for a comprehensive introduction of GP.

The \textit{measurements} presented in  Fig. \ref{pic_1} are the measured or numerically estimated state derivatives $\hat{f}$ based on measurements from the onboard sensors.
Based on the sensor data collected in real experiments, we assume that the uncertainty $d(s)$, which is the difference between the measurements $\hat{f}$ and the state derivatives $f_0$ predicted by the nominal model (\ref{eq:nominal}) under the same measured states and control inputs at each time step, is from sources discussed in Remarks 1 and 2. 
Its component in each member of the tracking states $s$ is independent and thus can be modeled as an independent Gaussian process.
For $N$ observations of the $j$th component of $d(s)$ at state points $S =[s_1,\dots,s_N]$, i.e. $\hat{d}^j=[\hat{d}^j_1,\dots,\hat{d}^j_N]^\top$, each observation is measured with independent Gaussian noise $\epsilon^j_i \sim\mathcal{N} (0, (\sigma^j_n )^2 )$, i.e. $\hat{d}^j_i=d^j(s_i)+\epsilon^j_i$.

A GP denoted by $\mathcal{GP}(\mu(s),k(s,s'))$ is defined with a set of hyperparameters $\theta$ chosen by a mean function $\mu(s)$ and a covariance kernel function $k(s,s')$ for capturing the characteristics of the uncertainty (linearity, periodicity, etc).
The prior distribution of $d^j$ modeled by a GP with an optimal $\theta$ is computed by maximizing the marginal likelihood of the observations. 
Induced by the prior distribution and observations $\hat{d}^j$, the posterior distribution of $d^j(s_\ast)$ at a new observation $s_\ast$ denoted by $\mathcal{N}(\Bar{d}^j(s_{\ast}), (\sigma^j(s_{\ast}))^2)$ is as follows:
\begin{equation}
    \begin{aligned}
      \Bar{d}^j(s_{\ast})=\mathbb{E}[d^j(s_{\ast})|\hat{d}^j,S] &= \mu^j(s_{\ast}) + K^j(s_{\ast},S)(K^j(S,S) + (\sigma^n_j )^2I)^{-1}(\hat{d}^j-\mu^j(S)),\\
      (\sigma^j(s_{\ast}))^2=\Sigma[d^j(s_{\ast})|S]&=K^j(s_{\ast},s_{\ast})- K^j(s_{\ast},S)((K^j(S,S) + (\sigma^n_j )^2I)^{-1}) K^j(S,s_{\ast}).
      \label{eq:GP_post}
    \end{aligned}
\end{equation}
The $j$th element of the \textit{uncertainty} model $\Tilde{d}(s, e)$ is constructed as 
\begin{equation}
    \Tilde{d}^j(s, e) = \Bar{d}^j(s)+ e\sigma^j(s),
    \label{eq:GP_uncertainy}
\end{equation}
where the uncertainty tuning variable $e\in\mathcal{E} = \{ \underbar{e}:||\underbar{e}||_{\infty} \leq \sqrt{2}\text{erf}^{-1}(p) \}$ is designed to provide chance-constrained safety guarantee, allowing that the probability of $d^j(s)$ lying within the bound $[\Bar{d}^j(s)-\sqrt{2}\text{erf}^{-1}(p)\sigma^j(s), \Bar{d}^j(s)+ \sqrt{2}\text{erf}^{-1}(p)\sigma^j(s)]$ is $p$, which is expected to close to 1, and erf($\cdot$) denotes the Gauss error function.

\subsection{The tracking, planning, relative models and the Hamilton-Jacobi reachability analysis}\label{sec_23}
The \textit{tracking} model $f = f_0 + \Tilde{d}(s,e)$, which is constructed as the combination of the nominal model introduced in section \ref{sec_21} and the uncertainty model estimated by GP in section \ref{sec_22}, is a less conservative estimation of the true USV dynamics compared to simply combining the nominal model with the worst-case bound $\mathcal{D}$ in \cite{b11}. 
Moreover, a simplified \textit{planning} model $h$ presented in the Fig.\ref{pic_1} with the planning states $p\in \mathcal{P}$ and the planning inputs $u_p\in \mathcal{U}_p$ is used for real-time trajectory planning. 
In this context, we use the 2D single integrator model
\begin{equation}
    \dot{p} = h(p,u_p)\triangleq
    \begin{bmatrix}
        \dot{x}_p\\\dot{y}_p
    \end{bmatrix}=
    \begin{bmatrix}
        \dot{u}_{x,p}\\\dot{u}_{x,p}
    \end{bmatrix}.
    \label{eq:planning2D}
\end{equation}

The \textit{relative} states $r\in\mathcal{R}$ of the relative model is derived by
\begin{equation}
    r = Q_ss-Q_pp, 
    \label{eq:relative_s}
\end{equation}
where $Q_s\in\mathbb{R}^{(5+n_{add})\times 5}$ adds $n_{add}$ additional tracking states, which are inputs of the uncertainty model (\ref{eq:GP_uncertainy}), to be the relative states, and $Q_p\in\mathbb{R}^{(5+n_{add})\times 2}$ matches the planning states $p$ and the relative states $r$.
Therefore, the relative states $r$ include the error between the tracking and planning models, the tracking states informing the evolution of these errors, and the additional tracking states that contribute to the uncertainty models. 
The \textit{relative} model is $\dot{r}=Q_s(f_0(s,u_s) + \Tilde{d}(s,e))-Q_ph(p,u_p)$.
In the case that the global $y$ position is the input of the uncertainty model $\Tilde{d}(s,e)$, the \textit{relative} model $\dot{r}= g(r, u_s, u_p, e)$ can be formulated as follow:
\begin{equation}
    \dot{r}= g(r, u_s, u_p, e)
    =    \begin{bmatrix}
        \Dot{x_r}\\
        \Dot{y_r}\\
        \Dot{\psi}\\
        \Dot{v}\\
        \Dot{\omega}\\
        \Dot{y}
    \end{bmatrix} = 
    \begin{bmatrix}
        v\cos{\psi} - u_{p,x} + \Bar{d}^1(s) + e_1 \sigma^1(s)\\
        v\sin{\psi} - u_{p,y} + \Bar{d}^2(s) + e_2 \sigma^2(s)\\
        \omega + \Bar{d}^3(s)+ e_3 \sigma^3(s)\\
        m^{-1}((n_1+n_2)F_{max}-(X_v+X_{|v|v}|v|)v) + \Bar{d}^4(s)+ e_4 \sigma^4(s)\\
        I^{-1}((n_1-n_2) \frac{L}{2} F_{max}-(N_{\omega}+N_{|\omega|\omega}|\omega|)\omega) + \Bar{d}^5(s)+ e_5 \sigma^5(s)\\
        v\sin{\psi} + \Bar{d}^2(s)+ e_2 \sigma^2(s)
    \end{bmatrix}
    \label{eq:relative_full}.
\end{equation}

With the relative model introduced, we present the second offline computation phase in Fig. \ref{pic_1}.
Following the FaSTrack framework \cite{b11}, we first define a cost function $l(r)$, i.e. $l(r) = ||\begin{bmatrix} x_r, y_r \end{bmatrix}^{\top}||_2$, as the cost for the pursuit-evasion game between the tracking and planning models. 
Next, we introduce the value function $V_{\infty}(r)$ to represent the highest cost the game will reach when both models are acting optimally.
In particular, it is defined as $V_{\infty}(r)\triangleq\lim_{T \rightarrow \infty} V(r, T)$, where
\begin{equation}
    V(r,T) = \sup_{\gamma_p\in\Gamma_p,\gamma_e\in\Gamma_e} \inf_{u_s(\cdot)\in\mathbb{U}_s} \left\{ \max_{\tau \in [0,T]} l\left( \xi_g \left( \tau;r,0,u_s(\cdot),\gamma_p[u_s](\cdot),\gamma_e[u_s](\cdot)  \right) \right) \right\}
    \label{eq:valueF}
\end{equation}
in which the tracking inputs are selected to minimize the cost, and the planning inputs/uncertainty tuning variable are selected to maximize the cost. 
Here, $\mathbb{U}_s$ represents the set of possible tracking inputs, $\Gamma_p$ and $\Gamma_e$ represent the set of mappings from tracking inputs to planning inputs/uncertainty tuning variable respectively, and $\xi_g$ is the trajectory of the relative model. 
These are described further in \cite{b11}.
By solving a Hamilton-Jacobi variational inequality from the value function and the relative model, 
\begin{equation}
    \max \left\{\frac{\partial V}{\partial t} + \min_{u_s\in\mathcal{U}_s}\max_{u_p\in\mathcal{U}_p, e\in\mathcal{E}}\left[\frac{\partial V}{\partial r}\cdot g(r,u_s,u_p,e)\right], l(r) - V(r,t)\right\} = 0,
    \label{eq:HJ_inequality}
\end{equation}
we can obtain the optimal controller, 
\begin{equation}
    u_s^*(r) = \arg \min_{u_s \in \mathcal{U}_s} \max_{u_p \in \mathcal{U}_p, \ e \in \mathcal{E}} \frac{\partial V(r)^{\top}}{\partial r} g(r,u_s,u_p,e),
    \label{eq:safe_control}
\end{equation}
which guarantees the relative model remains inside a robust positive set, called the TEB, given by $\mathcal{B} = \{ r| V_{\infty}(r) \leq \underbar{V}\}$, where $\underbar{V}= \max_{y \in \Xi_y} \min_{(x_r,y_r,\psi,v,\omega) \in \mathbb{R}^4} V_{\infty}(x_r,y_r,\psi,v,\omega,y)$, and $\Xi_y \subset R$ is the environment in the $y$ coordinate. 
We note this definition of $\underbar{V}$ is slightly different to standard FaSTrack \cite{b11} due to the dependence of the disturbance on $y$ and our need to ensure that the TEB exists at all possible $y \in \Xi_y$ for the purpose of obstacle augmentation during trajectory planning.

\subsection{Learning-based motion planning and control with safe guarantees}
\begin{algorithm}[h!]
\label{alg_1}
\caption{A framework for safety-guaranteed motion planning and control with GP estimation}\label{algorithm_1}
\begin{algorithmic}[1]

\STATE \textbf{OFFLINE STAGE} 
\STATE \textbf{Input:} Measurements $\hat{f}$, the uncertainty tuning variable $e$, the nominal model $f_0$, and the planning model $h$
\STATE Derive the estimated uncertainty model $\Tilde{d}(s, e)$ by following the steps in section \ref{sec_22}
\STATE Find TEB $\mathcal{B}$ and the safe controller $u_s^*(r)$ by following the steps in section \ref{sec_23}
\STATE \textbf{Output:} The TEB $\mathcal{B}$ and the safe controller $u_s^*(r)$
\STATE \textbf{ONLINE STAGE} 
\STATE \textbf{Input:} The TEB $\mathcal{B}$, the safe controller $u_s^*(r)$, a performance controller, a discretized planning model $h_d$, the obstacles $\mathbb{O}$, the goal $\mathbb{G}$, a reference $(x_{ref},y_{ref})$, and the initial tracking states $s_0$
\STATE Compute $\mathbb{O}_p$ from $\mathbb{O}$ and $\mathcal{B}_e(y)$ satisfying \eqref{eqn:aug-obst}
\STATE Find a feasible horizon T for the MPC problem in Algorithm 1 in [8] and the corresponding collision free trajectory by solving the MPC problem with $(x_0,y_0)$, $h_d$, $\mathbb{O}_p$, $\mathbb{G}$ and $(x_{ref},y_{ref})$
\STATE \textbf{For:} $k = 0,1,\hdots,T$
\STATE \quad Measure the tracking states $s(k)$ and evaluate the uncertainty model $\Tilde{d}(s(k),e)$
\STATE \quad \textbf{If:} The relative states $r(k)$ are inside interior of $\mathcal{B}$
\STATE \quad\quad Apply the performance controller
\STATE \quad \textbf{Else:}
\STATE \quad\quad Apply the safe controller $u^*_s(r_k)$
\end{algorithmic}
\end{algorithm}

The proposed learning-based trajectory planning and control framework is summarized in the \textbf{Algorithm 1}, which contains an offline stage and an online stage. 
The offline stage takes a set of measurements $\hat{f}$, the nominal model $f_0$, the planning model $h$ and the uncertainty tuning variable $e$ to compute the components of the trajectory planner and the hybrid controller in step 3 and 4, as presented in sections \ref{sec_22} and \ref{sec_23}

The trajectory planner and the hybrid controller can be combined with online applications to achieve safety-guaranteed trajectory planning and control.
Specifically, the role of the trajectory planner is to generate a trajectory for the USV to follow, such that when the hybrid controller is used to track the planned trajectory, collision avoidance is guaranteed. 
It takes in obstacles $\mathbb{O} \subset \mathbb{R} \times \Xi_y$ in the $x$-$y$ space, as well as the initial state $s_0\triangleq s(0)$ of the USV and a desired goal region $\mathbb{G} \subset \mathbb{R}^2$ in the $x$-$y$ space, and outputs a planned trajectory $p(k), \ k \in \{0, \hdots, T\}$, where $T$ is the number of time steps for trajectory planning. 
This is implemented by following the same strategy as \cite{b12} but modified for the time-invariant models we consider. 
Firstly, we define the $y$-dependent projected TEB $\mathcal{B}_{e}(y):=\{(x_r,y_r) \in \mathbb{R}^2| (\exists \psi,v,\omega \in \mathbb{R}) [ (x_r,y_r,\psi,v,\omega,y) \in \mathcal{B} ]\}$, and use it to compute the augmented obstacles in step 8:
\begin{equation}
    \mathbb{O}_p = \bigcup_{(x,y)\in\mathbb{O}}\left(\{ (x,y)\} \oplus (-\mathcal{B}_{e}(y)) \right). 
    \label{eqn:aug-obst}
\end{equation} 
\begin{remark}
In practice, the exact set $\mathbb{O}_p$ can be difficult to obtain, so an approximation of this set is computed by dividing the state space into a grid, and following a strategy like Algorithm 2 in \cite{b12}. 
The key difference is that our TEB is time-invariant, so we do not need to compute the augmented obstacle over every step of the horizon.
\end{remark}
Secondly, in step 9, the trajectory is planned by a solving a single open-loop MPC problem whose dynamics evolves according to an Euler discretized version of the 2D single integrator model in \eqref{eq:planning2D} (denoted by $h_d$), is initialised at $(x_0,y_0)$, avoids $\mathbb{O}_p$, reaches $\mathbb{G}$ and whose objective involves minimizing the distance between the planner states $p(k)$ and reference point $(x_{ref},y_{ref})\in \mathbb{G}$. 
Note that unlike \cite{b12}, we do not compute a diminished goal region in the planning space to avoid over-complicating the details, and thus goal-reaching is not guaranteed for the tracking model.
\begin{remark}
In Step 9 of Algorithm 1, we note that $T$ needs to be selected so that the MPC planning problem is feasible. 
One strategy for doing this is by solving the MPC problem with increasing horizon lengths until a feasible solution is obtained.
\end{remark}

The hybrid controller is responsible for tracking the planning model in a way such that collision avoidance is guaranteed, and good performance is achieved. 
These two objectives are achieved by allowing the hybrid controller to switch between two lower-level controllers; a performance controller, and a safe controller. 
The performance controller can be selected arbitrarily, but the safe controller is specifically chosen as $u_s^*(r)$. 
Using the invariance property from FaSTrack \cite{b11}, the safe controller guarantees that the tracking model will remain in a tube around the planning model position, described by $\mathcal{T}_e(x_p,y_p)=\{(x,y) \in \mathbb{R}^2| (\exists \psi,v,\omega\in \mathbb{R}) [ (x-x_p,y-y_p,\psi,v,\omega,y) \in \mathcal{B} ]\}$.
Then, so long as the relative model starts inside $\mathcal{B}$ and the planning model avoids the augmented obstacle $\mathbb{O}_p$, the tracking model is guaranteed to avoid the true obstacle. 
Safety is extended to the hybrid controller by carefully designing the switching method. 
In particular, if the relative states $r$ are within the interior of $\mathcal{B}$, the performance controller is applied. 
Otherwise, the safety controller is applied. We present the implementation of the hybrid controller in steps 10-16. 
\section{Simulation and experiments}
We first show in simulation a less conservative planned trajectory can be achieved with the GP-TEB, as compared to using the conservative TEB. 
Then, through experiments, we validate the framework and the chance-constrained safety guarantee in two separate tasks.

\subsection{Simulation: safety-guaranteed trajectory planning for USV} \label{sec_3.1}
\begin{figure}[h!]
 \centering
    \subfigure[]{\includegraphics[width=0.245\hsize]{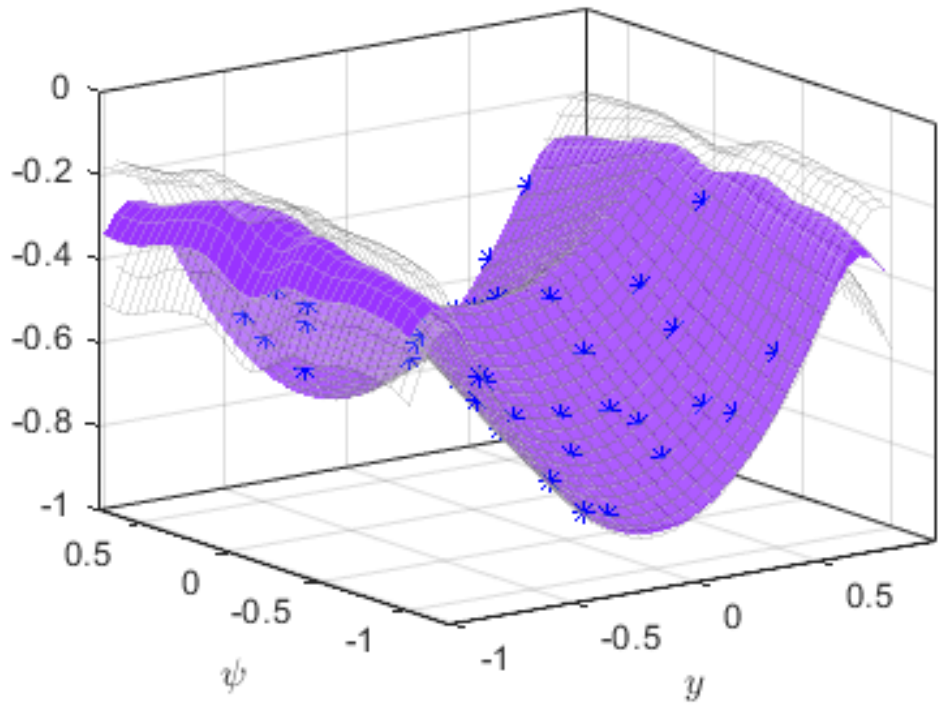}\label{gpvfig}}
    \subfigure[]{\includegraphics[width=0.24\hsize]{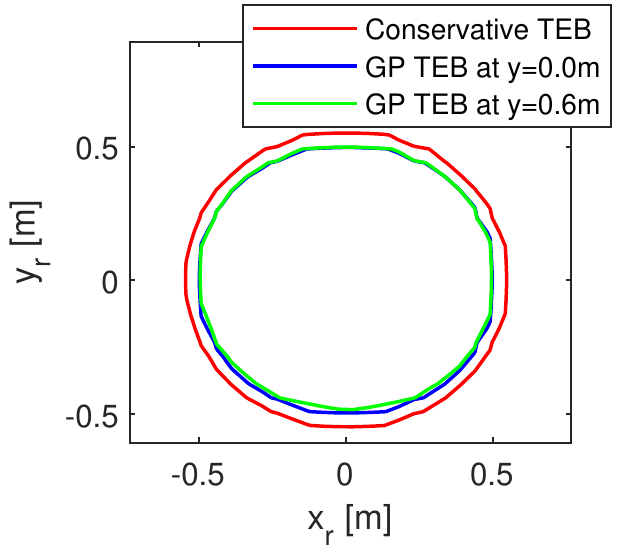}\label{tebfi}}
    \subfigure[]{\includegraphics[width=0.24\hsize]{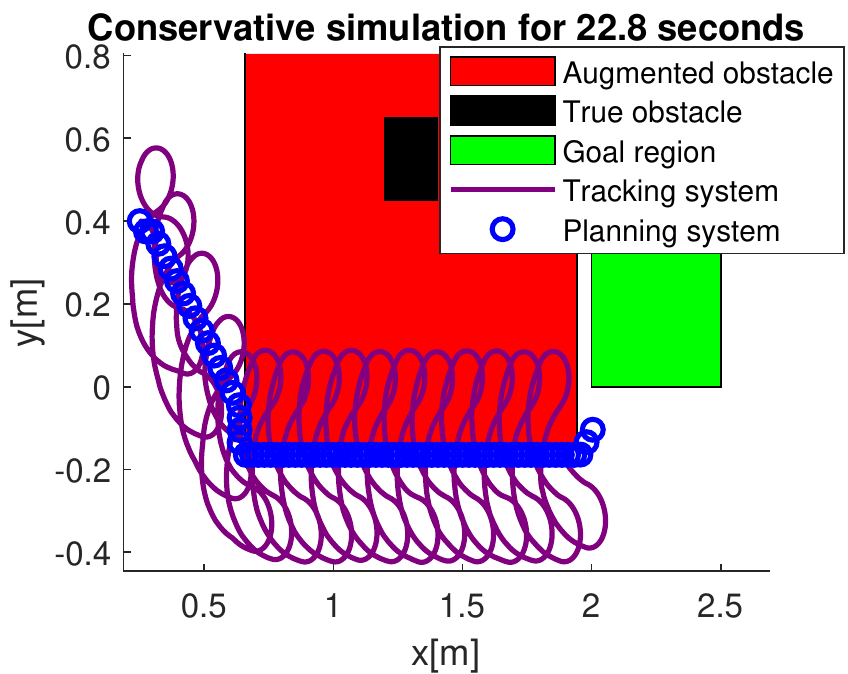}\label{gprfig4}}
    \subfigure[]{\includegraphics[width=0.245\hsize]{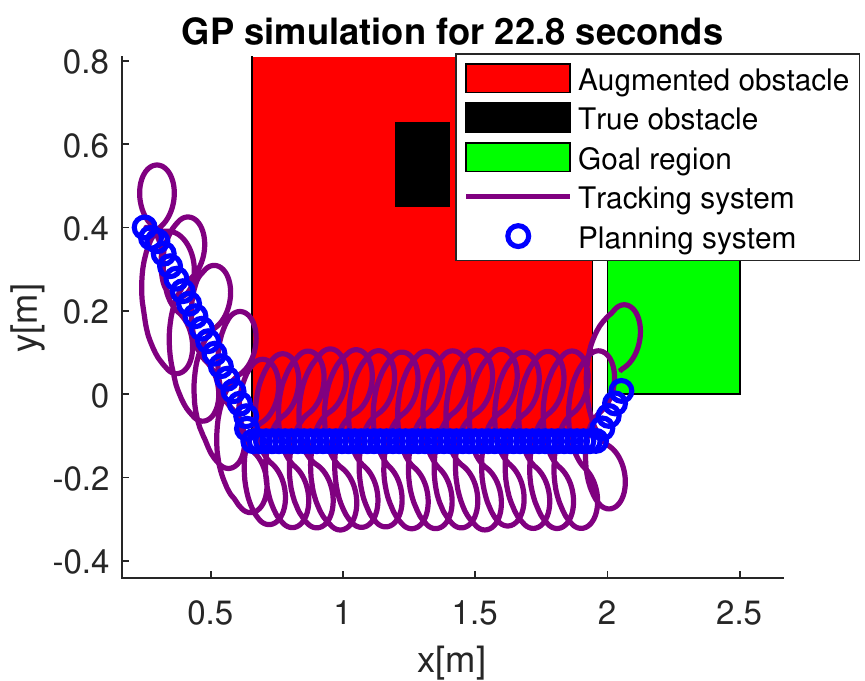}\label{gprfig5}}
\caption{(a) $\Tilde{d}_{\dot{x}}(s)$ with the mean function $\bar{d}_{\dot{x}}(s)$ (purple surface) and +/- 3 standard deviations $\sigma_{\dot{x}}(s)$ (transparent grids); (b) Conservative TEB vs. GP-TEB; (c) Conservative trajectory; (d) GP trajectory.}
\label{gp and teb}
\end{figure}

The USV is approximated by a 1-meter-diameter circle with an underlying unicycle model.
A handcrafted state-dependent disturbance given by 
$ d_{\dot{x}}(s) \triangleq  0.5(y^2-1)(1+\sqrt{\sin(\psi)^2})$
affects $\dot{x}$. Meanwhile, small bounded white-noise uncertainties affect $\dot{y}$ and $\dot{\psi}$. 
The simulation environment $\mathbb{O}$ constains a rectangular obstacle (black box at $x = 1.25$ meters), as shown in Fig. \ref{gprfig4}, \ref{gprfig5}.
The obstacle is pre-augmented by the USV's size so the USV can be considered a point mass during planning.
The USV needs to reach the goal region $\mathbb{G}$ (green box at $x = 2.25$ meters).

Fig. \ref{gpvfig} shows the disturbance $d_{\dot{x}}(s)$ learned by a GP model, following step 3 of Algo. 1 using observations (blue stars) uniformly sampled from $y \in [-1,1]$ and $\psi \in [-1,1]$. 
The learned model is given by $\Tilde{d}_{\dot{x}}(s, e)$, where the mean function is denoted as $\bar{d}_{\dot{x}}(s)$ (purple surface) and +/-3 standard deviations are denoted as $\sigma_{\dot{x}}(s)$ (transparent grids).
Based on $\Tilde{d}_{\dot{x}}(s, e)$ with +/-1 $\sigma_{\dot{x}}(s)$, the GP-TEB is calculated, whose projection onto the $x_r$-$y_r$-plane is shown in Fig. \ref{tebfi}.
Due to the state dependency of the GP-TEB, its size differs at $y=0$ and $y=0.6$. 
The conservative TEB is larger than the GP-TEB since it assumes the worst-case uncertainty bound $\mathcal{D}$ based on the largest value of the observations.

According to step 8 of Algo. 1, $\mathbb{O}$ is augmented by the TEB to produce $\mathbb{O}_p$, represented by the red box in Fig. \ref{gprfig4} and \ref{gprfig5}.  
Then, following step 9, a 22.8-seconds-long trajectory (sequence of blue circles in Fig. \ref{gprfig4} and \ref{gprfig5}) is planned for each case.
The safe controller $u^*_s(r)$ is applied always and the USV travels along the dark red path.
In both cases, $u^*_s(r)$ steers the USV away from the obstacle and maintains it in a neighborhood of the planned trajectory.
However, for the case with the conservative TEB, the trajectory planning problem becomes infeasible (unable to reach goal region). 

\subsection{Experiments using the Heron USV}
As shown in Fig. \ref{pic_2}, the heron USV is deployed in a 2.5-meters-wide water tunnel, compared to the USV's size of 1.35$\times$1 meters. 
The USV relies on a ZED2i tracking camera in conjunction with an extended Kalman filter to obtain observations for training as well as provide new measurements for online control.
A laptop with a MATLAB-ROS interface sends real-time control commands from the hybrid controller to the USV.
The experiments were carried out at low speeds ($v \leq 0.5$ m/s) and in windless weather to prevent excessive disturbances.
The USV is again approximated by a 1-meter-diameter circle.
The conservative TEB is not considered because of infeasibility in planning.

\subsubsection{Experiment 1: safe control for center line tracking}
In this task, the USV attempts to stay near the centerline of the tunnel, captured by setting $y_p=0$, and $x$ is not concerned.
The USV's initial state $s_0 = (y=0.5,\psi=0,v=0,\omega=0)$.
The performance controller is kept constant at $(\tau_1 = 0.17, \tau_2 = 0.19)$, inducing forward and left-turning motion of the USV. 
The disturbances corresponding to $\dot{\omega}$ and $\dot{v}$ are captured by GP models with +/-1 standard deviation, based on which the safe controller $u^*_s(r_k)$ and GP-TEB are derived.
Since no planning stage is required, steps 10-15 of Algo. 1 is directly applied to control the USV.

\begin{figure}[ht!]
 \centering
    \subfigure[]{\includegraphics[width=0.245\hsize]{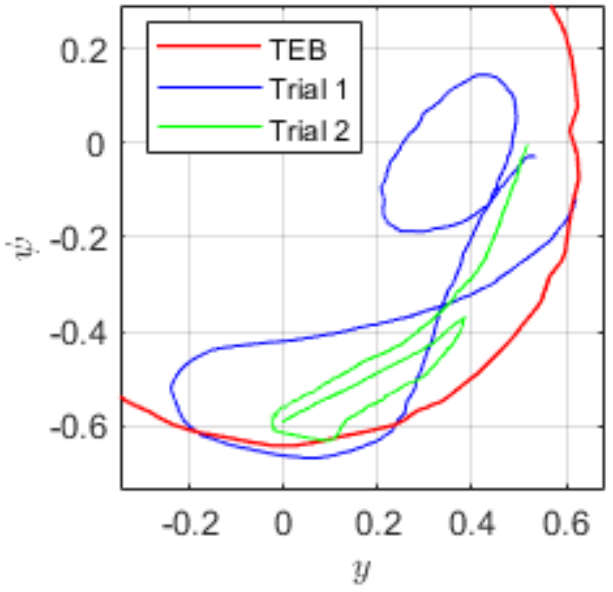}\label{trial1}}
    \subfigure[]{\includegraphics[width=0.24\hsize]{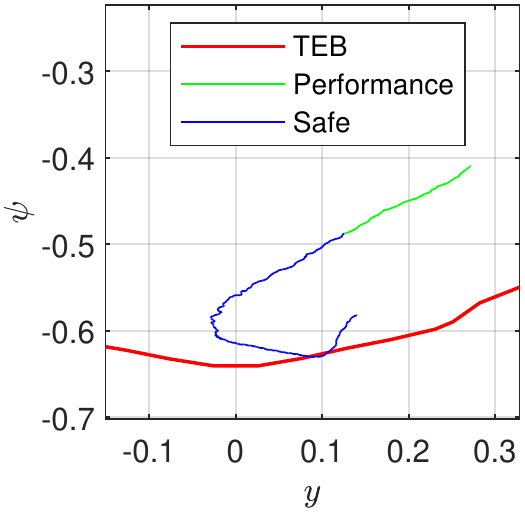}\label{trial2}}
    \subfigure[]{\includegraphics[width=0.245\hsize]{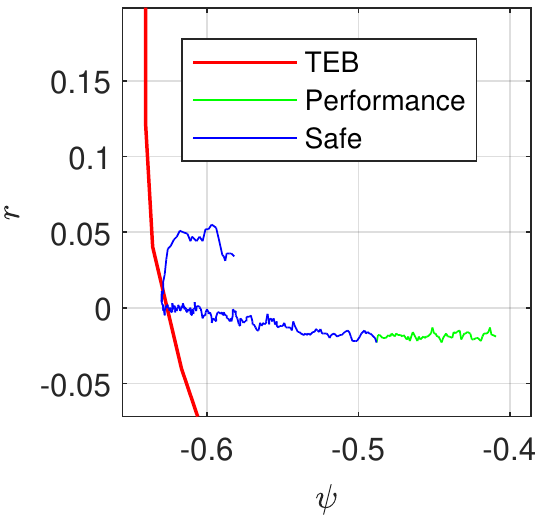}\label{trial3}}
    \subfigure[]{\includegraphics[width=0.24\hsize]{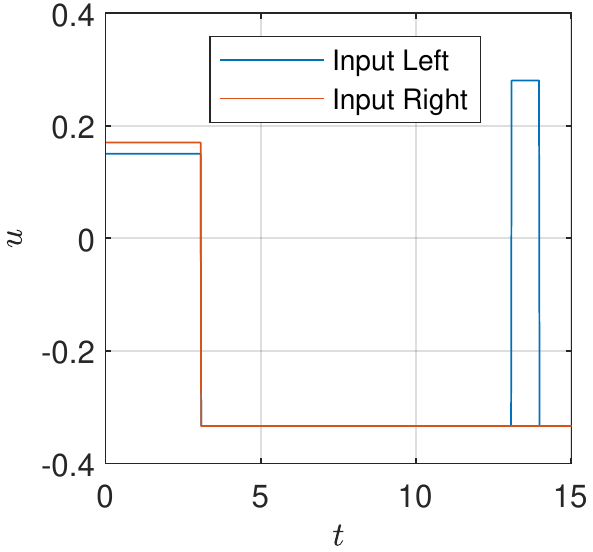}\label{trial4}}
\caption{(a) Trajectories of two selected trials; (b)(c) Partial trajectory of trial 2; (d) Input of trial 2. }
\label{global error upper bound}
\end{figure}

Fig. \ref{trial1} shows the USV's state trajectories in two selected trials in the $y$-$\psi$-plane, and the corresponding projected TEB.
The state trajectories mostly stay within the TEB, while trial 1 temporarily escapes the TEB, which is expected because of the +/- 1 standard deviation chosen.
Fig. \ref{trial2}, \ref{trial3}, and \ref{trial4} gives a close look at trial 2, where the USV was initially safe with the performance controller applied. Then, as $r$ approaches the border of TEB, $u^*_s(r)$ was triggered to slow the USV down and then drive the USV clockwise to prevent it from escaping the TEB in the $\psi$-axis.

\subsubsection{Experiment 2: safety-guaranteed motion planning and control (Alg. 1) for a USV} 
\begin{figure}[h!]
\centering  
\includegraphics[width=0.77\hsize]{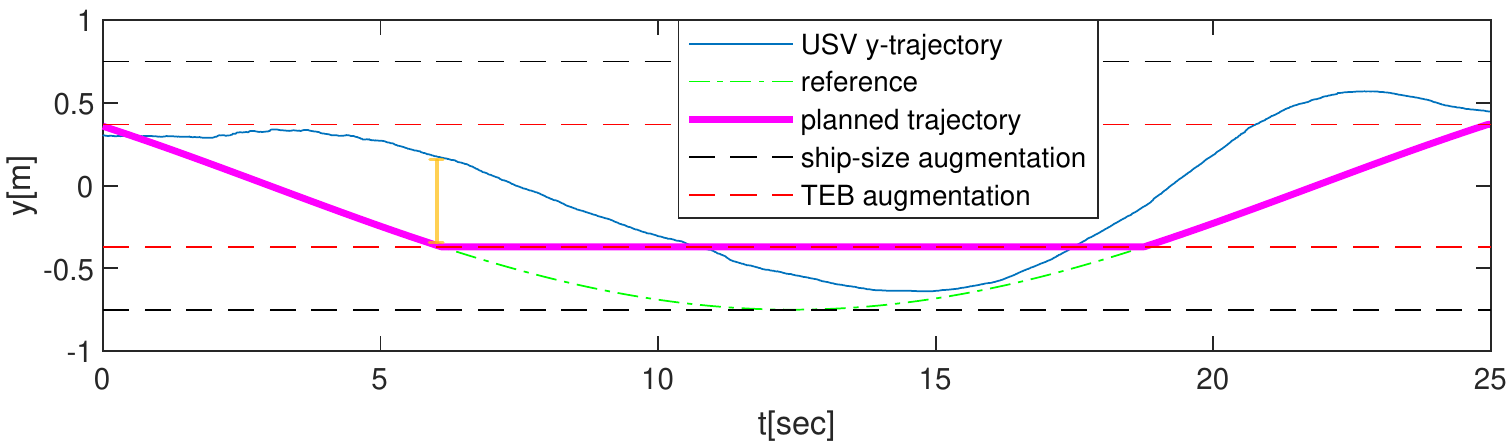}
\caption{Trajectory tracking with the USV}
\label{exp2 trial1}
\end{figure}

In this task, the USV travels at $v=0.35$ m/s constantly and tracks a sinusoidal reference trajectory in $y$ while avoiding collision with the tunnel.
Safely tracking such a trajectory enables the USV in missions such as continued water quality monitoring or patrolling. 
As shown in Fig. \ref{exp2 trial1}, $s_0 = (y_r=y=0.35,\psi=0,\omega = 0)$.
The tunnel is first augmented by the size of the USV  to produce $\mathbb{O}$ (between the black dashed lines), where the reference trajectory (green dashed curve) is designed to stay inside.
The tunnel is further augmented by the GP-TEB with +/-1 standard deviations (red dashed lines) to produce $\mathbb{O}_p$.
Step 9 of Algo. 1 is used to produce the planned trajectory (magenta curve).
Due to the already limited space and uncertainties, only the safe controller $u^*_s(r)$ is used and the USV's state changes according to the $y$-trajectory (blue curve).
It can be seen that the USV is able to stay within a close neighborhood of the planned trajectory and avoid collision with the augmented edges of the tunnel (black dashed lines).

\clearpage


\clearpage
\begin{thebibliography}{00}
\bibitem{b10} T. I. Fossen, “Maneuvering Theory,” in \textit{Handbook of Marine Craft Hydrodynamics and Motion Control}. Hoboken, NJ, USA: Wiley, 2011.

\bibitem{b1_seth} T. Lolla, P. F. Lermusiaux, M. P. Ueckermann, and P. J. Haley, “Time-optimal path planning in dynamic flows using level set equations: Theory and schemes,” \textit{Ocean Dynamics}, vol. 64, no. 10, pp. 1373–1397, 2014. 

\bibitem{b2_seth} N. Yang, D. Chang, M. R. Amini, M. Johnson-Robersor and J. Sun, "Energy Management for Autonomous Underwater Vehicles using Economic Model Predictive Control," \textit{2019 American Control Conference (ACC)}, 2019, pp. 2639-2644, doi: 10.23919/ACC.2019.8815106.

\bibitem{b3_seth} D. Q. Mayne, E. C. Kerrigan, E. J. van Wyk, and P. Falugi, “Tube-based robust nonlinear model predictive control,” \textit{International Journal of Robust and Nonlinear Control}, vol. 21, no. 11, pp. 1341–1353, 2011. 

\bibitem{b4_seth} A. Majumdar and R. Tedrake, “Funnel libraries for real-time robust feedback motion planning,” \textit{The International Journal of Robotics Research}, vol. 36, no. 8, pp. 947–982, 2017. 

\bibitem{b5_seth} X. Xu, P. Tabuada, J W. Grizzle, A D. Ames, “Robustness of Control Barrier Functions for Safety Critical Control,” in \textit{AHDS}, 2015

\bibitem{b11} M. Chen, S. Herbert, H. Hu, Y. Pu, J. Fernandez Fisac, S. Bansal, S. Han, and C. J. Tomlin, “Fastrack: A modular framework for real-time motion planning and Guaranteed Safe Tracking,” \textit{IEEE Transactions on Automatic Control}, pp. 1–1, 2021. 

\bibitem{b12} S. Siriya, M. Bui, A. Shriraman, M. Chen and Y. Pu, "Safety-Guaranteed Real-Time Trajectory Planning for Underwater Vehicles in Plane-Progressive Waves," \textit{2020 59th IEEE Conference on Decision and Control (CDC)}, 2020, pp. 5249-5254.

\bibitem{b13} J. F. Fisac, A. K. Akametalu, M. N. Zeilinger, S. Kaynama, J. Gillula, and C. J. Tomlin, “A General Safety Framework for learning-based control in uncertain robotic systems,” \textit{IEEE Transactions on Automatic Control}, vol. 64, no. 7, pp. 2737–2752, 2019. 

\bibitem{b14} Clearpath Robotics Inc., “Heron - Génération Robots.” [Online]. Available: https://www.generationrobots.com/media/clearpath\_heron\_usermanual.pdf. [Accessed: 04-Oct-2021]. 

\bibitem{b24} A. K. Akametalu, J. F. Fisac, J. H. Gillula, S. Kaynama, M. N. Zeilinger, and C. J. Tomlin, “Reachability-based safe learning with Gaussian Processes,” \textit{53rd IEEE Conference on Decision and Control}, 2014.

\bibitem{b22} N. Manzini, “USV path planning using potential field model,” \textit{M.S.M.E Thesis}, Dept. of Mech. and Aero. Eng., NPS, Monterey, CA, USA, 2017. [Online]. Available: http://hdl.handle.net/10945/56152

\bibitem{b23} C. E. Rasmussen and C. K. I. Williams, \textit{Gaussian processes for machine learning}, volume 14. MIT Press, April 2006.



\end{thebibliography}
\end{document}